# Robust Stochastic Parsing Using the Inside-Outside Algorithm

**Ted Briscoe & Nick Waegner**

(ejb@cl.cam.ac.uk)

University of Cambridge, Computer Laboratory

Pembroke Street, Cambridge, CB2 3QG, UK

## 1 Introduction

Development of a robust syntactic parser capable of returning the unique, correct and syntactically determinate analysis for arbitrary naturally-occurring input will require solutions to two critical problems with most, if not all, current wide-coverage parsing systems; namely, resolution of structural ambiguity and undergeneration. Typically, resolution of syntactic ambiguity has been conceived as the problem of representing and deploying non-syntactic (semantic, pragmatic, phonological) knowledge. However, this approach has not proved fruitful so far except for small and simple domains and even in these cases remains labour intensive. In addition, some naturally-occurring sentences will not be correctly analysed (or analysed at all) by a parser deploying a generative grammar based on the assumption that the grammatical sentences of a natural language constitute a well-formed set (e.g. Sampson, 1987a,b; Taylor *et al.*, 1989). Little attention has been devoted to this latter problem; however, the increasing quantities of machine-readable text requiring linguistic classification both for purposes of research and information retrieval, make it increasingly topical. In this paper, we discuss the application of the Viterbi algorithm and the Baum-Welch algorithm (in wide use for speech recognition) to the parsing problem and describe a recent experiment designed to produce a simple, robust, probabilistic parser which selects an appropriate analysis frequently enough to be useful and deals effectively with the problem of undergeneration. We focus on the application of these stochastic algorithms here because, although other statistically based approaches have been proposed (e.g. Sampson *et al.*, 1989; Garside & Leech, 1985; Magerman & Marcus, 1991a,b), these appear most promising as they are computationally-tractable (in principle) and well-integrated with formal language / automata theory.

The Viterbi algorithm and Baum-Welch algorithm are optimised algorithms (with polynomial computational complexity) which can be used in conjunction with stochastic regular grammars (finite-state automata, i.e. (hidden) markov models, Baum, 1972) and with probabilistic context-free grammars (Baker, 1982; Fujisaki *et al.*, 1989) to select the most probable analysis of a sentence and to (re-)estimate the probabilities of the rules (non-zero parameters) defined by the grammar (respectively). The Viterbi algorithm computes the maximally probable derivation with polynomial resources despite the exponential space of possible derivations (e.g. Church & Patil, 1983) by pruning all non-maximal paths leading to the set of non-terminals compatible with the input at each step in the parsing process. The Baum-Welch algorithm (which is often called the inside-outside algorithm with context-free grammars) computes the probability of each possible derivation with polynomial resources also by factoring the computation across each non-terminal involved in any derivation. A detailed and clear description of these algorithms is provided by Charniak (1993), Holmes (1988) and Lari & Young (1990), amongst others. These algorithms will converge towards a local optimum when used to iteratively re-estimate probabilities on a training corpus in a manner which maximises the likelihood of the training corpus given the grammar.

It is possible to imagine exploiting these algorithms in a number of ways in text processing and parsing and, so far, relatively few of the possible options have been explored. To date the primary application of these techniques in text rather than speech processing has been the use of the Viterbi algorithm in the lexical tagging of corpora with part-of-speech categories, training on an unambiguous corpus (e.g. de Rose, 1988). Typically a tagged training corpus is used to train a bigram or trigram (first- or second- order) ergodic finite-state machine (i.e. no parameters are set to zero, which is equivalent to assuming that no grammatical constraints are assumed other than those imposed by the choice of tagset). This represents one of the simplest applications of such techniques, because the unambiguous training data ensures that the model will converge to the true optimum and the ergodic assumption ensures that all possible derivations will involve the same number of states for any given length of input. Recently, Cutting *et al.* (1992) have developed a tagging system based on the Baum-Welch algorithm, trained on untagged data which performs as well as these Viterbi based systems.

In what follows we will only consider the application of these algorithms to probabilistic context-free grammars (PCFGs) and extensions of such models, since we are addressing problems of parsing rather than tagging.

## 2 Choosing Between Analyses

Fujisaki *et al.* (1989) describe a corpus parsing experiment using a PCFG containing 2118 rules which was first converted into Chomsky Normal Form (CNF) (creating 7550 productions) and then trained on an ambiguous corpus of 4206 sentences using a variant of the Baum-Welch re-estimation procedure. In this case the model was constrained in the sense that many of the possible parameters (rules) defined over the category set were set to zero before training began. Thus training was used only to estimate new probabilities for a set of predefined rules. The utility of the resulting probabilities was evaluated by testing the trained grammar on sentences randomly selected from the training corpus, using the Viterbi algorithm to select the most probable analysis. In 72 out of 84 sentences examined, the most probable analysis was also the correct analysis. 6 of the remainder were false positives and did not receive a correct parse, whilst the other 6 did but it was not the most probable. A success rate (per sentence) of 85% is apparently impressive, but it is difficult to evaluate properly in the absence of further details concerning the nature of the corpus. For example, if the corpus contains many simple and similar constructions, training on ambiguous data is more likely to converge quickly on a useful set of probabilities. (Fujisaki *et al.* report that the majority of their corpus had an average sentence length of 10.85 words.)

Sharman *et al.* (1990) conducted a similar experiment with a grammar in ID/LP format. ID/LP grammars separate the two types of information encoded in CF rules — immediate dominance and immediate precedence — into two rule types which together define (a subset of) the CFLs. This allows probabilities concerning dominance, associated with ID rules, to be factored out from those concerning precedence, associated with LP rules. In this experiment, an unambiguous training corpus of sentences paired with a semantically appropriate syntactic analysis was employed consisting of about 1 million words of text. A grammar containing 100 terminals and 16 non-terminals and initial probabilities based on the frequency of ID and LP relations was extracted from the training corpus. The resulting probabilistic ID/LP grammar was used to parse 42 sentences of 30 words or less drawn from the same corpus. In addition, lexical tag probabilities were integrated with the probability of the ID/LP relations to rank parses. 18 of the most probable parses (derived using a variant of the Viterbi algorithm) were identical to the original manual analyses, whilst a further 19 were 'similar', yielding a success rate of 88%. What is noticeable about this experiment is that the results are not significantly better than Fujisaki *et al.*'s experiment with ambiguous training data discussed above, despite the use of more unambiguous training data, a more sophisticated language model and a grammar derived directly from the corpus (thus ruling out undergeneration). It seems likely that these differences derive from the differential complexity of the corpus material used for training and testing, properties of the grammars employed, and so forth. However, these two results underline the need for the use of shared corpora in training and testing, or model / grammar independent measures of complexity, such as estimation of the actual entropy / perplexity of a language (Sharman, 1990; Wright, 1990).

Briscoe & Carroll (1993) and Carroll (1993) extend the approach to probabilistic generalised LALR(1) parsers. The motivation for this move is that LR parse tables (which are non-deterministic finite-state machines in the generalised case) provide a more fine-grained probabilistic model than PCFGs, and thus can distinguish probabilistically derivations involving (re)application of identical grammatical rules (such as in typical analyses of noun-noun compounding or PP attachment) in different orders. They construct a LALR(1) parse table from the CF backbone of the Alvey Natural Language Tools (ANLT) grammar, a wide-coverage unification-based grammar (e.g. Briscoe *et al.*, 1987) and derive a probabilistic version of the table by interactively guiding the LR parser to the semantically appropriate analysis of the training data (i.e. an unambiguous training corpus is semi-automatically created using the parser / grammar to be trained). The resulting LR parse histories are used to associate probabilities with the table directly (in contrast to Wright (1990) and others, who have proposed to 'compile' the probabilities associated with a PCFG into the LR table). A packed parse forest representation is constructed with probabilities associated with (sub)analyses in the forest and the parse forest is unpacked to recover the the n-best parses (Wright *et al.*, 1991; Carroll, 1993). This approach was tested on a corpus of noun definitions taken from the *Longman Dictionary of Contemporary English* (LDOCE) training on 151 definitions and testing on 63 definitions from the training data and a further 54 unseen definitions. The definitions vary in length from 2 to 31 words and in global ambiguity from unambiguous to over 2500 distinct analyses. The most probable parse was correct in around 75% of cases for both unseen and seen data. In the case of the unseen data, most of the failures are false positives in which no correct analysis is found and in the remaining cases the correct analysis is most frequently one of the three most probable analyses. In the case of the false positives, the most frequent cause of failure was the lack of a subcategorisation frame in the set of lexical entries associated with a word.

Each of these experiments suggests that probabilistic modelling may be useful in selecting an appropriate parse from the set licensed by a generative grammar. However, none achieve results reliable enough to be of practical utility and none address the problem of undergeneration or grammar induction via stochastic techniques in a manner analogous to work in speech recognition or lexical tagging.

## 3 Dealing with Undergeneration

The techniques presented above whilst being useful for the disambiguation of analyses, are as 'brittle' as standard approaches when presented with new examples, for which the correct analysis cannot be assigned, or in the limit, for which no analysis is possible. These situations may arise, through a deficiency in the syntactic rules or lexicon, or simply where the input is ill-formed or extragrammatical. One approach to this problem, is to iteratively develop the grammar, adding suplementary rules, and re-analysing the failed examples (e.g. Black *et al.* 1992). The aim here is to ensure broad coverage by the labour intensive analysis of as large a subset as possible of the target language. In more lexically-orientated approaches to grammar, it can be expected that the principle cause of undergeneration will be the incompleteness of lexical entries (Briscoe & Carroll, 1993). However, manual correction or development of a realistic lexicons does not appear feasible, given the vast amount of coding required (Boguraev & Briscoe 1989), and there are good reasons to believe that the goal of developing an entirely watertight generative grammar is unattainable (Sampson, 1987a; Taylor *et al.*, 1989).

A potential solution to the problem of undergeneration using the inside-outside algorithm is suggested by the work of Lari & Young (1990). They utilise a tabular parsing algorithm (e.g. CYK) coupled with a PCFG in CNF. Initially, they assume that all possible CNF rules which can be formed from a prespecified terminal and non-terminal category set are possible; that is, are associated with non-zero probabilities. The inside-outside algorithm is used to re-estimate the probabilities of these rules by repeated iteration over a corpus until they stabilise within some prespecified threshold. In this way a (locally) optimal set of rules and probabilities is induced which maximise the probability of the language defined by the corpus. Thus they propose an approach which is the CF counterpart of making the ergodic assumption in (hidden) markov modelling.

One problem with this technique is that the search space of possible parses even for small category sets is very large. But, although this is defined by the the number of binary branching trees over a sentence of length $n$ (the Catalan series, Church & Patil, 1983) multiplied by the number of possible labellings of the nodes in that tree (the number of non-terminals to the power of $n$-1), exploitation of an efficient parse forest representation reduces the computation complexity of the algorithm to $n^3 \times G^2$, where $n$ is the length of the input and where G is the number of grammar rules. Nevertheless, the algorithm is only practical for small (non-terminal) category sets. Lari & Young generated a corpus of 200 random palindromes from a grammar containing 5 non-terminals, two terminals and 8 rules (non-zero parameters) for the simple palindrome language $\{xy|x$ is a mirror image of $y\}$. The same category set with all 135 possible rules (parameters) given initial non-zero probabilities was used to re-estimate the grammar. After approximately 70 iterations the system stabilised on a weakly-equivalent grammar for this language. Lari & Young (1990) demonstrate that this grammar is a better model of the observed language than that produced by a hidden markov model with the same number of parameters, in the sense that the predicted entropy of the language is lower. Unsurprisingly, the CFG is also better able to classify members of the palindrome language since this language cannot be generated using a regular grammar.

It is fairly clear that scaling up Lari & Young's approach for a realistic grammar of natural language, such as the ANLT grammar, would be computationally intractable — the ANLT CF backbone contains 575 distinct categories (before conversion to CNF). It might be thought that switching to the Viterbi algorithm would allow a considerable saving in computation since the latter only requires computation of the average and maximum probabilities for each sentence of the corpus for re-estimation. However, in order to achieve convergence it would almost certainly be necessary to use much more training data in this case, because less information is being extracted from each example. It would probably be necessary to use a smoothing technique to avoid unseen events converging towards zero too rapidly. And, it is impossible to combine filtering of CF backbone parses by unifying the remaining features with the Viterbi algorithm, since this can mean that the most probable CF backbone analysis is invalid, necessitating backtracking through sub-optimal analysis paths.

One simplification would be to use lexical tagging of corpora to reduce the size of the terminal vocabulary and to make it unambiguous. It might then be possible to develop a grammar of the order of complexity of that typical used in the manual parsing of tagged corpora, such as the Lancaster skeleton parsing scheme (Leech & Garside, 1991) or the Penn Treebank scheme (Santorini, 1990). These schemes typically assume of the order of 10 non-terminal symbols; however, even grammars of this simplicity result in startlingly large search spaces when no grammatical constraints (other than CNF) are assumed; for instance, for a 20 word sentence and 10 non-terminals over a determinate lexical input there are $1.76726319 \times 10^{28}$ possible analyses and for a 30 word sentence $1.00224221665137 \times 10^{44}$. Although it is possible to perform re-estimation in parallel quite straightforwardly by splitting the corpus, these figures suggest that this approach may be too computationally expensive to be practical.

Furthermore, Pereira and Schabes (1992) show that the grammars acquired through re-estimation from unbracketed tag sequences do not express the constituency assigned by linguists, but rather tend to bracket tags with high mutual dependencies. To ameliorate this problem they utilise bracketed training data and only re-estimate from derivations in which there are no crossing brackets between the input and the analysis assigned. In this way, rules which create unconventional constituency are weeded out during re-estimation. However, the use of bracketed training data derived from treebanks limits the utility of the approach, as treebanking is a labour-intensive process (e.g. Leech and Garside, 1991). Furthermore, although the resultant PCFG produces unlabelled bracketings which coincide to a promising extent with those assigned manually, the non-terminal labels associated by the trained grammar with these constituents bear no relation to linguistically-meaningful categories. It might be possible to relabel them, but there is no guarantee that the trained grammar will assign a unique label to, say, noun phrases in this approach.

In order to explore whether it is possible to utilise Baum-Welch re-estimation with a more linguistically plausible grammar and train from unbracketed tag sequences, we developed a variant of the approach in which an explicit grammar is created by hand and a set of implicit rules is automatically added to this grammar to extend coverage. Baum-Welch re-estimation is then utilised to discover which of the implicit rules are more useful and to refine the initial probability distribution, which is high for explicit and low for implicit rules. Some of the advantages od this approach are that no manual parsing is required for training, the resulting labelled analyses are linguistically comprehensible, the technique is not restricted to CNF CFGs, or indeed to CFGs, and it is possible, in principle, to utilise grammars containing larger non-terminal category sets because the increase in grammar size when implicit rules are added is reduced as it is constrained by grammatical constraints.

## 4 Imposing Grammatical Constraints

In the limit, imposing grammatical constraints on the initial model used for re-estimation would reintroduce the problem of undergeneration and reduce Lari & Young's technique into one for the acquisition of probabilities from an ambiguous corpus for a completely pre-specified grammar (as with the experiments described in section 2). However, it is possible to envisage an intermediate position in which some broad grammatical constraints are imposed and some rules are explicitly specified with higher initial probabilities, whilst implicit rules compatible with these constraints are assigned a floor non-zero probability, and illegal rules incompatible with the constraints are not considered. In this way, the search space defined over the category set can be reduced and the size of the category set can be increased, whilst the initial bias of the system will be towards a linguistically motivated (local) optimum. In what follows, we suggest several constraints and propose a general feature-based approach to their specification. The idea is that many constraints will have the property of ruling out linguistically uninterpretable analyses without necessarily constraining weak generative capacity.

### 4.1 Headedness

The notion of headedness as expressed, for example, in X-bar Theory (e.g. Jackendoff, 1977), can be formalised in a feature-based unification grammar containing rules of the type illustrated in Figure 1, which is specified in the ANLT formalism (e.g. Briscoe et al., 1987).

If we think of the word declarations as a set of unary rules, rewriting preterminals as terminals, and the aliased categories as atomic category symbols, G1 specifies a CNF CFG with 11 non-terminal and 20 terminal categories (given in full in appendix 1). If we were to induce a CNF PCFG from G1 using Lari & Young's technique, by assigning a floor probability to all possible rules, this would force the following (and many other) rules to be considered and thus incorporated into possible analyses assigned by the parser:

```
FEATURE V{+, -}
FEATURE BAR{0, 1, 2}

ALIAS V2 = [V +, N -, BAR 2].
ALIAS V1 = [V +, N -, BAR 1].

PSRULE S1 : V2 --> N1 V1.
PSRULE VP1 : V1 --> V0 N1.

WORD cat : N0.
WORD the : DT.
```

Figure 1: Simple X-bar Grammar Rules

```
N2 --> V2 V2        N2 --> P0 V1
A1 --> V2 N1        A1 --> P0
V2 --> N2 P1        V2 --> A0 A1
```

Linguistically, these rules are unmotivated and implausible for the same reason: they violate the constraint that a phrase must have a head; for example, a noun phrase (N2) must contain a noun (N), a sentence (V2) must contain a verb (phrase) (V1), and so forth. Of course, there are many more possible combinations of the category set for G1 which also violate this constraint and taken together they can be used to define a very large number of possible analyses of input sentences. Furthermore, the interpretation of such rules within any extant linguistic framework is impossible, so it is unclear what we would 'learn' if the system converged on them. However, if we impose the following constraint on formation of further rules in G1, then all of the above rules will be blocked:

```
CONSTRAINT HEAD1 :
    [N, V, BAR(NOT 0)] --> [], [];
    N(0)=N(1), V(0)=V(1),
    BAR(0)=(BAR(1) | BAR(1) -- 1).
```

This (meta)grammatical constraint consists of a rule pattern (up to the semicolon) and a constraint on feature values (expressed using identical syntax to propagation rules in the ANLT formalism; Briscoe et al., 1987). It is to be interpreted as a constraint on possible immediate dominance relations in CF rules consisting of a rule schema specifying that all rules must contain a non-(pre)terminal mother category and two non-terminal daughters, that all mother categories must be specified for N, V and BAR, and that one daughter must share these values or have a BAR value of one less than the mother. Thus this constraint also blocks rules containing heads with BAR values higher than that of the mother category. The utility of such rules is also dubious since they express linguistically implausible claims, such as the head of a phrase is a clause, and so forth. The constraint licenses rules not in G1 such as:

```
a)                  b)
V2 --> N2 V2        V2 --> V1 V1
```

```
V2 --> A1 V2        V2 --> P0 A1
V1 --> V0 V1        V1 --> V1 V2
N1 --> N0 V2        N1 --> N0 N1
```

The four rules in a) constitute linguistically motivated extensions to G1 but those in b) are harder to justify, indicating that, although 'headedness' can provide useful restrictions on possible rules, it is not the whole story. For convenience, in the first experiment we impose the further constraint given below, which restricts rules introducing two preterminals so that the second daughter must always be A0 or N0:

```
CONSTRAINT PT1 :
   [] --> [BAR 0] [BAR 0];
                  N(2)= +.
```

This constraint interacts with HEAD1 to define a further 99 implicit rules not in G1. Many of these rules are linguistically unmotivated. Nevertheless, it may be that the X-bar schema does provide enough constraint, taken together with the CNF constraint and an initial probabilistic bias in favour of the original rules, to make the approach practical and useful. The number of parameters (explicit and implicit rules) in the probabilistic model defined by G1's constraints is of the same order as that used by Lari & Young for the palindrome language. Thus the space of possible analyses remains of similar size, although it is much reduced over the space defined using G1's category set and allowing any possible CNF rule; for instance, for the 14 word sentence *passionately with the sheep the cat chases the ball with the boy so slowly* the implicit grammar provides approximately 380,000 analyses (as V2), whilst the number Catalan(14) $\times$ $11^{14}$ is considerably bigger.

The main motivation for using the inside-outside algorithm and raising the floor of implicit rules is to be able to parse unexpected orderings of terminal categories. Implicit G1 does not quite allow any possible ordering of terminal categories — any sentence ending with an adverb not proceeded by a degree modifier cannot be parsed, for instance. Nevertheless, we can demonstrate that with respect to G1 and one very pervasive form of word order that we will not prevent the system finding a linguistically motivated analysis. The explicit portion of G1 can analyse a) below without modification.

```
a) A girl kisses a boy so passionately
b) A girl so passionately kisses a boy
c) So passionately a girl kisses a boy
d) ? A girl kisses so passionately a boy
e) * A so passionately girl kisses a boy
```

However, b) and c) require the addition of the two implicit rules below:

```
V1 --> A1 V1              V2 --> A1 V2
```

The analysis of d), which is an unlikely but possible example in a stylistically marked context, or of e) which is plain ungrammatical would require the addition of the rules in a) and b) below, respectively:

```
a) V1 --> V0 A1    V1 --> V1 N2
b) N1 --> A1 N1
```

Of course, there are other possible ways, using implicit G1, of parsing these examples and it is an empirical question whether the system will stabilise on these analyses.

### 4.2 Experiment 1

We generated 500 sentences using a probabilistic version of explicit G1 (probabilities used are given in brackets after the rules in appendix 1). We then produced a probabilistic version of implicit G1 with the implicit rules given a floor probability of around 0.01 and the explicit rules initialised with higher probability. This gave a grammar with a total of 126 CNF CF rules (27 of which were explicit rules derived from G1 PS rules and word declarations). As a simple test we trained this grammar using the inside-outside algorithm[1] on the 500 sentences. It was then retrained on an larger corpus, consisting of the original 500 sentences, and 28 hand-written examples which could only be analysed with the addition of the implicit rules, such as *slowly with the sheep the boy chases the ball*. In this example, an adverb occurs without a degree modifier and both the adverbial and prepositional phrases are preposed. Explicit G1 does not contain rules covering these possibilities. The resulting trained grammar is given in appendix 1 (with rules with zero probability excluded). In Figure 2 we give two measures of the entropy (per word) (Wright, 1990) of the source language and the estimated language from the original 500 sentences. For comparison we provide the same measures for the palindrome language investigated by Lari & Young (1990). In the case of the implicit grammar trained on the corpus of 528 sentences, only the entropy of the randomly initialised grammar and the trained grammar are shown, as the entropy of the source language is unknown. These measures are defined by:

$$H^{3a} = -\frac{\sum_K \log P(S)}{\sum_K |S|}$$

$$H^{3b} = -\frac{1}{K} \sum_K \frac{\log P(S)}{|S|}$$

where $P(S)$ and $|S|$ are the probability and length of sentence $S$ respectively and $K$, the number of sentences in the set. For the extended corpus we show the difference in entropy between the implicit grammar with random probability assignment and the trained grammar. These figures demonstrate that the inside-outside algorithm is converging on a good optimum for modelling both the original language and the extended language given the initial conditions. They also demonstrate that the language being modelled has an entropy roughly twice that of the palindrome language. Thus we have shown that restricting the space of possible rules that is explored and biasing initial parameters towards a linguistically motivated optimum allows the model to converge very rapidly to a useful solution. In this case, the model converges after only 6 iterations over the training corpus, suggesting that we may be able to extend the approach successfully to more complex grammars / languages. Crucially, these results extend to the case where the original explicit grammar is only an approximate (undegenerating) model of the training data. This situation recreates (in a small way) the situation in which the linguistically-motivated grammar available undergenerates with respect to naturally-occurring sentences.

---
[1] The version of the inside-outside algorithm used throughout this paper is that presented in Jelinek (1985)

| Entropy Measure | $H^{3a}$ | $H^{3b}$ |
|---|---|---|
| Palindrome Language | 0.6870 | 0.7266 |
| Estimated Panlindrome Lang. | 0.6916 | 0.7504 |
| Explicit Grammar 500 | 1.5954 | 1.5688 |
| Implicit Grammar 500 | 1.5922 | 1.5690 |
| Initial Grammar 528 | 2.0979 | 2.0898 |
| Trained Grammar 528 | 1.5584 | 1.5698 |

Figure 2: Measures of Entropy

Although these results are promising, we are crucially interested in the analyses assigned by the trained grammar and not in its ability to model the language (strings). One measure of success here is the extent to which the trained grammar has zeroed the parameters for implicit rules. In the final trained version of implicit G1, 52 non-zero rules remain (27 explicit rules + 25 implicit rules). Recall that we said that there are approximately 380,000 analyses for the 14 word sentence *passionately with the sheep the cat chases the ball with the boy so slowly*; this example has 75 parses in the trained grammar. In addition, we analysed 14 sentences parsed using the trained grammar, recording the most probable analysis assigned, the probability of this analysis, the total number of analyses, the probability of all analyses and the likelihood of the analysed sentence given the trained grammar. These sentences are drawn from the original 500, the additional 28 and further unseen examples. Examination of the parse trees shows that the trained grammar is not perfect, in the sense that not all the constituents constructed conform to linguistic intuitions; for example, the constituent [N1 [A0 passionately A0] [N1 [DT the DT] [N0 boy N0]N1]N1]. In addition, global ambiguities such as PP attachments are not resolved in a manner which necessarily accords with our semantic intuitions. Nevertheless, the system has done about as well as could be expected given only information about rule frequencies. Furthermore, in the cases where the examples are 'nearly' grammatical, in the sense that they deviate from the explicit grammar by no more than one or two rules, the analyses assigned are almost always the 'closest fit' that can be achieved using a minimal number of implicit rules. In many cases, this results in the linguistically-motivated analysis being induced. The most ambiguous example (17 words long) has 273 parses, the average is just under 60 parses (for average length of 13.5 words). Ignoring PP attachment ambiguity, 8 rules are misapplied out of a total of 160 rule applications for these examples, yielding a figure of 95% correct rule application for the examples analysed.

## 5 Feature-based Encoding of Constraints

In most implementations of X-Bar Theory a feature-based encoding of headedness is assumed and, at least since GPSG (Gazdar *et al.*, 1985), the feature theory is formalised via the unification operation (e.g. Shieber, 1984). Within a broad framework of this type it is possible to envisage imposing many grammatical constraints by treating feature-based generalisations as constraints on the 'compilation' of a (CF) phrase structure grammar (PSG). For example, we could add an agreement constraint to G1 by requiring, for instance, that daughters in rules which have agreement features NUM and PER the values of these features must be consistent, as in AGR below:

```
CONSTRAINT AGR :
    [] --> [NUM, PER], [NUM, PER];
                NUM(1)=NUM(2),
                PER(1)=PER(2).
```

This rule blocks the generation of PS rules in which the values of these features differ or, if variable, are not bound. We have extended the ANLT metagrammatical grammar compilation system to incorporate this approach to grammatical constraints or partial grammar specification. In the current ANLT grammar compiler, these rules are used to create a set of 'fleshed out' PS rules in which aliases are expanded out, constraints applied in the form of feature variable or value bindings, and so forth. There are two ways that the compiled out grammar might be interfaced to a system, such as that of Lari & Young (1990), which assumes a CNF CFG.

Firstly, we might expand out PER, NUM and any other features with variable values, creating new rules for all possible values of these features according to the feature declarations. This approach is guaranteed to terminate if feature declarations specify a finite number of values, and the result will be a set of categories which can trivially be replaced by a new set of atomic symbols. However, in general this approach is likely to lead to impractical increases in the size of the grammar for the purposes of tractable re-estimation of probabilities using the inside-outside algorithm. It also means that feature-based grammatical constraints can only be employed in a manner which allows compilation into a CFG, precluding the use of category-valued features and other linguistically common techniques which lead to an in-finite feature system / category set. Secondly, we can simply re-alias the non-variable parts of categories using the existing aliases (for perspicuity) and filter with the remaining features for the purposes of assigning parses to sentences during the first phase of re-estimation. Alternatively, Briscoe & Carroll (1993) provide an algorithm which automatically constructs the most informative CF backbone from a unification-based grammar in which categories are specified entirely in terms of feature sets. Unification of features could be treated as either a 'hard' constraint to remove certain analyses from the re-estimation process or possibly in a 'softer' fashion to adjust probabilities in a manner sensitive to this phase of the parse process.

### 5.1 Experiment 2

The availability of large quantities of tagged and hand-corrected corpora, such as the Penn Treebank, LOB, SEC, Susanne and others, coupled with the relative reliability of automatic tagging (e.g. de Rose, 1988), means that an obvious test (and potential useful application) for a robust parser would be in the automatic parsing of tag sequences to construct analyses of the same order of complexity as those currently constructed manually (see above). Most tagged corpora contain between 50-120 distinct lexical tags. These tags most often encode PER and NUM information as well as major category information. We can, therefore, create a lexicon of tags in which each tag is represented as a feature set with determinate values for all features:

```
NNS : [N +, V -, BAR 0, PER 3, NUM Sg].
NNP : [N +, V -, BAR 0, PER 3, NUM Pl]. etc.
```

We have developed a unification-based grammar (G2) for the CLAWS2 tagset (Garside et al., 1987:Appendix B) containing 156 lexical categories (tags), 17 features (maximum number of values 15), 8 non-terminal (aliased) categories, 12 terminal (aliased) categories, 104 binary-branching PS rule (schemata), and 10 constraints of feature propagation and defaulting (of the type described above). These constraints implement headedness, agreement, and also constrain the grammar of coordination via the propagation and defaulting of a feature CONJ. Implicit rules are automatically generated from G2 by creating a new set of CF rules encoding all possible binary rules from the set of aliased categories defined in G2. Then category declarations are used to expand out the aliased categories in these potential implicit rules with their featural definitions. The constraints of G2 are applied to produce bindings and default values in these rules. Any potential implicit rule which does not match the pattern specified by a propagation constraint is filtered out of the set of implicit rules.

We have used G2 to produce explicit and implicit CNF CFGs for use with the inside-outside re-estimation and parsing system. However, were all the features in the PS rules to be expanded out to create a CNF CFG, the resulting explicit grammar would consist of 63,831 rules. Combining these with the implicit rules licensed by the constraints in G2, would generate over 250,000 rules, which is too many for our current implementation and hardware, and would also lead to problems of data insufficiency. We chose instead to re-alias a subset of the features in the set of rules produced and form a CNF CFG from these aliases. In this way, we can control the size of the category set to keep the re-estimation technique tractable. Thus, we simplified the grammar by not utilising featural distinctions between sub-classes of the major categories, thereby yielding a total of 8271 rules (1850 of which were explicit). This simplification both increased slightly the coverage of the grammar and so too the number of spurious analyses assigned to any given tag sequence. Whilst the simplified explicit grammar parses less than 20% of the SEC, the combined grammar (consisting of both implicit and explicit rules) can assign a complete parse to about 75% of the corpus.

Once again the rules were initialised randomly prior to training, with explicit rules initialised with higher probabilities than implicit rules. After 5 iterations of the inside-outside algorithm, during which very low probability rules were set to zero, 3786 rules with non-zero probability remained. Using this trained grammar, 14 sentences selected at random from the corpus were analysed, of which 10 were assigned complete parses. The Viterbi algorithm was used to extract the most probable parse, together with its probability and the number of explicit rules employed. Using the inside phase of the inside-outside algorithm, the probability of all analyses of the sentence, the number of analyses and the likelihood of the most probable parse were calculated. Appendix 3 contains a number of these analyses, with a brief comment on the errors associated with the most probable analysis. As can be seen from these examples, approximately 90% of rules used in the most probable parse were explicit rules. This is only to be expected, as these rules are assigned higher probabilities at initialisation. However, it also demonstrates that typically only a few extra rules are necessary in order to modify the grammar to increase coverage. Concerning the errors in the most probable parses: although the grammar is extremely ambiguous, the most probable parse in 8 out of 10 of the complete parses is close to correct, and in the case of observation 7 completely correct. Comparing the most probable analyses with the syntactico-semantic most plausible bracketing and major category assignment yields a correct rule application rate of 79% (39 errors out of 189 applications). Given that about a third of these errors concern level of attachment of arguments / modifiers, requiring semantic disambiguation, these results suggest that the technique is promising.

## 6 Further Experiments

Waegner (1993) extended these experiments by applying the approach described above to a larger and different corpus, applying a consensus evaluation methodology, and extending the technique to non-CNF CFGs and to unification-based grammars.

For the experiments summarised below the training corpora consisted of tagged unpunctuated sentences (max. length 30 words) extracted from the manually parsed treebanks of the SEC and the Associated Press

(AP) corpus of newspaper articles treebank developed jointly by IBM and Lancaster University. 2461 sentences (average length 13.4 words) were extracted from the SEC and 2528 sentences (average length 17.3 words) from AP. A further 121 sentences from SEC (average length 18.6 words) 225 sentences from AP (average length 17 words) were extracted for evaluation. In this summary, results for both corpora are merged; in general, results for SEC were slightly worse, perhaps reflecting the fact that it is transcribed spoken language (see Waegner, 1993 for further details).

The evaluation scheme used is that proposed by the Grammar Evaluation Interest Group (GEIG, see Black, Abney et al., 1991). It involves comparing the unlabelled bracketing derived from the most probable analysis yielded by the grammar/parser against that extracted from the manually parsed treebanks. The comparison is expressed in terms of recall, precision and crossing of brackets. Recall is defined as the percentage of brackets present in the treebank also present in the automatic parse. Precision is the percentage of brackets present in the automatic parse also present in the treebank. Crossings occur when an automatic/manual bracketing pair contains at least one word in common but neither is a subset of the other. Crossings measure the degree to which the automatic parse is in clear conflict with the treebank. Recall measures the degree to which the automatic parse agress with the treebank. Precision measures the degree to which the automatic parse 'extends' the treebank analysis – as treebanks tend to have flatter structure and analysts tend to leave difficult material unbracketed (e.g. Leech and Garside, 1991), it is not clear whether such extensions are correct or incorrect.

The framework was extended from CYK parsing of CNF CFGs to bottom up active chart parsing of arbitrary CFGs and the grammatical constraints were extended to allow the definition (or induction) of ternary branching rules (for example for verb complementation rules) without licensing the generation of infinite numbers of implicit rules. A Viterbi-like algorithm was used to extract the most probable analysis from the chart. These extensions allowed the use of a more linguistically conventional grammar which better reflected conventional assumptions about constituency.

### 6.1 CFG explicit/implicit grammar

The first experiment consisted of re-estimation of the new explicit PCFG and of the explicit/implicit PCFG from initial probabilities, assigned randomly for the explicit grammar and as above for the explicit/implicit grammar. After six iterations of inside-outside re-estimation, the explicit grammar converged to 979 rules above the baseline threshold. After five iterations, the explicit/implicit grammar converged to 3789 rules above this threshold. The explicit grammar covered 56.2% of the training sentences and the explicit/implicit grammar 93.5% of them. Comparisons were made between the most probable analyses drawn from initial and trained grammars with the treebank analyses of the test sentences using the GEIG scheme. The results are shown in Figures 3 and 4.

| Grammar | Initial | Trained |
|---|---|---|
| Sentences Parsed (No. / %) | 181 / 54.35 | 180 / 54.05 |
| Average Sentence Length | 15.90 | 15.90 |
| Total Recall (%) | 58.30 | 66.30 |
| Total Precision (%) | 40.52 | 45.64 |
| Average Crossings | 16.07 | 12.73 |

Figure 3: Results for Explicit Grammar

| Grammar | Initial | Trained |
|---|---|---|
| Sentences Parsed (No. / %) | 320 / 96.1% | 319 / 95.8% |
| Average Sentence Length | 17.67 | 17.67 |
| Total Recall (%) | 50.07 | 62.51 |
| Total Precision (%) | 36.78 | 42.57 |
| Average Crossings | 21.80 | 15.16 |

Figure 4: Results for Explicit/Implicit Grammar

The explicit/implicit technique is quite successful at dealing with undergeneration, since coverage improves to around 96% on both training and test sentences. (The slight decline in coverage for trained grammars results from one explicit rule being discarded since no exemplar of the relevant construction occurs in the training data.) Comparison of the precision, recall and crossings figures for the random and trained grammars indicates that, although there is a small degradation of the accuracy of the analyses yielded by the explicit/implicit grammar, the increase in coverage does not render the grammar significantly less accurate in terms of parse selection (especially as the increase in sentence length will increase the average ambiguity of the test sentences). However, it is also clear that whilst coverage might be said to have reached practical levels, the trained grammars will not deliver the most likely analysis reliably. It is probable that the performance revealed by the GEIG scheme reflects a similar level of accuracy as that shown by the manual evaluation of results in section 5.1.

### 6.2 Unification-based explicit/implicit grammars

In a second experiment, Waegner (1993) tried reincorporating the featural constraints excluded in the construction of a manageable PCFG for the experiments above. The grammar was modified so that the PCFG formed a backbone and each application of a rule involved unification of the residue of features in the manner specified in the original feature-based grammar. In the case of unification failure, the relevant derivation was discarded during re-estimation of the probabilities of backbone rules. Thus, a partially probabilistic unification-based grammatical model was developed.

In principle, incorporating these constraints should serve to make re-estimation more accurate by blocking consideration of spurious ambiguities. The explicit grammar in this form parsed 38.5% of the training corpus and after four iterations 898 rules remained. The explicit/implicit grammar parsed 89.2% and after four iterations 4274 rules remained. The reduction in coverage over the earlier experiments is due to the greater stringency (and inaccuracy) of the unification-based constraints. Despite the reduction in training data, the results for the trained grammars in Figures 5 and 6 demonstrate a marginal improvement in parse accuracy over those for pure PCFG, suggesting that this approach is worth further exploration.

| Sentences Parsed (No. / %) | 177 / 53.15 |
|---|---|
| Average Sentence Length | 15.81 |
| Total Recall (%) | 66.66 |
| Total Precision (%) | 46.05 |
| Total Crossings | 2441 |
| Average Crossings | 13.79 |

Figure 5: Explicit Grammar with Unification

| Sentences Parsed (No. / %) | 319 / 95.80 |
|---|---|
| Average Sentence Length | 17.67 |
| Total Recall (%) | 63.69 |
| Total Precision (%) | 43.34 |
| Total Crossings | 2441 |
| Average Crossings | 15.25 |

Figure 6: Explicit/Implicit Grammar with Unification

## 7 Conclusions

The experiments reported above demonstrate that Baum-Welch re-estimation combined with linguistically-motivated constraints on implicit rule generation is a powerful technique for extending coverage of PCFGs and of PCFG backbones to unification-based grammars. However, the results also demonstrate that parse selection using PCFGs does not lead to systems with good performance. This is not surprising given the inadequacies of PCFG for natural language (e.g. Briscoe and Carroll, 1993; Magerman and Marcus, 1991b). Future work must focus on combining these techniques for controlled rule induction with more context-dependent models of parse selection.

## Acknowledgements

We would like to thank John Carroll, Fernando Pereira and Steve Young for their help and advice. Any errors in this paper remain our responsibility.

# Appendix 1 — G1: A simple X-bar Grammar

```
FEATURE N{+, -}                              WORD cat : N0.             (0.15)
FEATURE V{+, -}                              WORD bird : N0.            (0.2)
FEATURE BAR{0, 1, 2}                         WORD park : N0.            (0.1)
FEATURE MINOR{DT, DG}                        WORD ball : N0.            (0.2)
                                             WORD girl : N0.            (0.08)
ALIAS V2 = [V +, N -, BAR 2].                WORD boy : N0.             (0.15)
ALIAS V1 = [V +, N -, BAR 1].                WORD sheep : N0.           (0.12)
ALIAS V0 = [V +, N -, BAR 0].                WORD chases : V0.          (0.65)
ALIAS N1 = [V -, N +, BAR 1].                WORD kisses : V0.          (0.35)
ALIAS N0 = [V -, N +, BAR 0].                WORD in : P0.              (0.4)
ALIAS P1 = [V -, N -, BAR 1].                WORD with : P0.            (0.6)
ALIAS P0 = [V -, N -, BAR 0].                WORD slowly : A0.          (0.72)
ALIAS A1 = [V +, N +, BAR 1].                WORD passionately : A0.    (0.28)
ALIAS A0 = [V +, N +, BAR 0].                WORD the : DT              (0.4)
ALIAS DT = [MINOR DT].                       WORD a : DT.               (0.3)
ALIAS DG = [MINOR DG].                       WORD this : DT.            (0.1)
                                             WORD that : DT.            (0.3)
PSRULE S1 : V2 --> N1 V1.   (1.0)            WORD so : DG.              (0.3)
PSRULE VP1 : V1 --> V0 N1.  (0.9)            WORD too : DG.             (0.25)
PSRULE VP2 : V1 --> V1 A1.  (0.1)            WORD very : DG.            (0.45)
PSRULE NP1 : N1 --> DT N0.  (0.8)
PSRULE N1 : N1 --> N1 P1.   (0.2)
PSRULE P1 : P1 --> P0 N1.   (1.0)
PSRULE A1 : A1 --> DT A0.   (1.0)
```

# Appendix 2 — Probabilistic CNF Trained Version of Implicit G1

```
V2   -->   V2        P1        0.00057531 implicit
V2   -->   V2        A1        0.00076667 implicit
V2   -->   N1        V1        0.94625349 explicit
V2   -->   N0        V1        0.00541748 implicit
V2   -->   P1        V2        0.00693598 implicit
V2   -->   A1        V2        0.02500444 implicit
V2   -->   A0        V2        0.01504663 implicit
V1   -->   V1        P1        0.00050115 implicit
V1   -->   V1        A1        0.07906031 explicit
V1   -->   V0        N1        0.90583286 explicit
V1   -->   V0        P1        0.00171885 implicit
V1   -->   P1        V1        0.00606044 implicit
V1   -->   A1        V1        0.00166985 implicit
V1   -->   A0        V1        0.00515654 implicit
V0   -->   chases    #         0.71401515 explicit
V0   -->   kisses    #         0.28598485 explicit
N1   -->   N1        P1        0.17184879 explicit
N1   -->   N1        A1        0.00003595 implicit
N1   -->   N0        A1        0.00060919 implicit
N1   -->   P1        N1        0.00074166 implicit
N1   -->   A1        N1        0.00085517 implicit
N1   -->   A0        N1        0.00166976 implicit
N1   -->   DT        N0        0.82423948 explicit
N0   -->   cat       #         0.16212233 explicit
N0   -->   bird      #         0.19528371 explicit
N0   -->   park      #         0.09874724 explicit
N0   -->   ball      #         0.21075903 explicit
N0   -->   girl      #         0.08032424 explicit
N0   -->   boy       #         0.15401621 explicit
N0   -->   sheep     #         0.09874724 explicit
```

```
P1    -->    V1             P1          0.00328333 implicit
P1    -->    P1             P1          0.00099618 implicit
P1    -->    P0             N1          0.99571773 explicit
P1    -->    A0             P1          0.00000276 implicit
P0    -->    in             #           0.44554455 explicit
P0    -->    with           #           0.55445545 explicit
A1    -->    V1             A1          0.00001139 implicit
A1    -->    P1             A1          0.03045650 implicit
A1    -->    A1             P1          0.00592188 implicit
A1    -->    A1             A1          0.03028750 implicit
A1    -->    A0             P1          0.11100389 implicit
A1    -->    A0             A1          0.00120497 implicit
A1    -->    DG             A0          0.82111386 explicit
A0    -->    slowly         #           0.66250000 explicit
A0    -->    passionately # 0.33750000 explicit
DG    -->    so             #           0.27586207 explicit
DG    -->    too            #           0.27586207 explicit
DG    -->    very           #           0.44827586 explicit
DT    -->    the            #           0.43754619 explicit
DT    -->    a              #           0.29120473 explicit
DT    -->    this           #           0.09460458 explicit
DT    -->    that           #           0.17664449 explicit
```

## Appendix 3 — SEC Parses with G2

A) Complete Parses
------------------------------------------------------------------------
Observation 1

Next_MD week_NNT1 a_AT1 delegation_NN1 of_IO nine_MC Protestant_JJ
ministers_NNS2 from_II Argentina_NP1 visits_VVZ the_AT Autumn_NN1
assembly_NN1 of_IO the_AT British_JJ Council_NNJ of_IO Churches_NNJ2

Parsed sentence:
```
[V2.2
  [N2.2
   [N1.4 [A0.1 Next_MD A0.1][N0.2 week_NNT1 N0.2]N1.4]
   [N2.2
    [DT.4 a_AT1 DT.4]
    [N1.4 [N1.4 [N0.2 delegation_NN1 N0.2]
                [P1.2 [P0.7 of_IO P0.7]
                      [N2.7
                       [DT.3 nine_MC DT.3]
                       [N1.2 [A1.1 Protestant_JJ A1.1]
                             [N0.1 ministers_NNS2 N0.1]N1.2]N2.7]P1.2]N1.4]
          [P1.2 [P0.7 from_II P0.7]
                [N2.2 Argentina_NP1 N2.2]P1.2]N1.4]N2.2]N2.2]
  [V1.2 [V0.13 visits_VVZ V0.13]
        [N2.2 [DT.4 the_AT DT.4]
              [N1.4 [N0.2 [N0.2 Autumn_NN1 N0.2][N0.2 assembly_NN1 N0.2]N0.2]
                    [P1.2 [P0.7 of_IO P0.7]
                          [N2.2 [DT.4 the_AT DT.4]
                                [N1.4 [A1.1 British_JJ A1.1]
                                      [N1.4 [N0.2 Council_NNJ N0.2]
                                            [P1.2 [P0.5 of_IO P0.5]
                                                  [N2.1 Churches_NNJ2 N2.1]
                                      P1.2]N1.4]N1.4]N2.2]P1.2]N1.4]N2.2]V1.2]V2.2]
```

Sentence length: 19 words
best 5.809595e-33 all 1.064649e-30 likelihood   0.005457

(Tot number of parses : 21862499278031036 )

Total Rules Applied 40 Total Explicit 36 (90.00%)
Ratio of correct rules / rules applied: 18/19

Comments: 'Next week' not part of N2 but V2; 'from Argentina' attach lower?

------------------------------------------------------------------------
Observation 2

More_DAR news_NN1 about_II the_AT Reverend_NNS1 Sun_NP1 Myung_NP1
Moon_NP1 founder_NN1 of_IO the_AT Unification_NN1 church_NN1 who_PNQS
's_VBZ currently_RR in_II jail_NN1 for_IF tax_NN1 evasion_NN1

Parsed sentence:
[V2.2
  [N2.2
   [N1.4
    [A1.2
     [A0.2 More_DAR A0.2]
     [N1.4 [N0.2 news_NN1 N0.2]
           [P1.2 [P0.7 about_II P0.7]
                 [N2.2
                   [DT.4 the_AT DT.4]
                   [N1.4
                    [N0.2
                     [N0.2
                      [N0.2
                       [N0.2 [N0.2 Reverend_NNS1 N0.2] [N0.2 Sun_NP1 N0.2]N0.2]
                       [N0.2 Myung_NP1 N0.2]N0.2]
                      [N0.2 Moon_NP1 N0.2]N0.2]
                     [N0.2 founder_NN1 N0.2]N0.2]
                    [P1.2 [P0.7 of_IO P0.7]
                          [N2.2 [DT.4 the_AT DT.4]
                                [N0.2 Unification_NN1 N0.2]N2.2]P1.2
                    ]N1.4]N2.2]P1.2]N1.4]A1.2]
    [N0.2 church_NN1 N0.2]N1.4]
   [N2.4 who_PNQS N2.4] N2.2]
  [V1.2 [V0.3 's_VBZ V0.3]
    [P2.1 [A1.4 currently_RR A1.4]
          [P1.2 [P0.7 in_II P0.7]
                [N2.2 [N1.4 [N0.2 jail_NN1 N0.2]
                            [P1.2 [P0.8 for_IF P0.8]
                                  [N0.2 tax_NN1 N0.2]P1.2]N1.4]
                      [N0.2 evasion_NN1 N0.2]N2.2]P1.2]P2.1]V1.2]V2.2]

Sentence length: 21 words
best 8.557940e-33 all 1.232113e-29 likelihood   0.000695
(Tot number of parses : 31241634778345856 )

Total Rules Applied 42  Total Explicit Rules Applied 38 ( 90.48 % )
Ratio of correct rules / rules applied: 14/20

Comments: 'more' not head; 'founder' not part of name; 'unif church'
split; 'currently' not in PP, 'tax evasion' split; N2 --> N1 N0 = too
probable implicit rule; relative clause split

------------------------------------------------------------------------
Observation 3

he_PPHS1 was_VBDZ awarded_VVN an_AT1 honorary_JJ degree_NN1 last_MD

```
week_NNT1 by_II the_AT Roman_JJ Catholic_JJ University_NNL1 of_IO
la_&FW Plata_NP1 in_II Buenos_NP1 Aires_NP1 Argentina_NP1

Parsed sentence:
[V2.2
 [N2.2 he_PPHS1 N2.2]
 [V1.2
  [V1.2 [V0.3 was_VBDZ V0.3]
   [V1.1 [V0.12 awarded_VVN V0.12]
    [N2.2 [DT.4 an_AT1 DT.4]
         [N1.4 [A1.1 honorary_JJ A1.1][N0.2 degree_NN1 N0.2]N1.4]N2.2]V1.1]V1.2]
  [N2.4 [N1.4 [A1.1 [A0.1 last_MD A0.1]
              [N1.4 [N0.2 week_NNT1 N0.2]
                    [P1.2 [P0.7 by_II P0.7]
                          [N2.2
                            [DT.4 the_AT DT.4]
                            [N1.4 [A1.1 Roman_JJ A1.1]
                              [N1.4
                                [A1.1 Catholic_JJ A1.1]
                                [N1.4 [N0.2 University_NNL1 N0.2]
                                      [P1.2 [P0.7 of_IO P0.7]
                                            [N2.2 la_&FW N2.2]P1.2]N1.4]N1.4]
                                                   N2.2]P1.2]N1.4]A1.1]
                              [N1.4 [N0.2 Plata_NP1 N0.2]
                                    [P1.2 [P0.7 in_II P0.7]
                                          [N2.2 Buenos_NP1 N2.2]P1.2]N1.4]N1.4]
          [N0.2 [N0.2 Aires_NP1 N0.2][N0.2 Argentina_NP1 N0.2] N0.2]
                                                   N2.4]V1.2]V2.2]

Sentence length: 20 words
best 1.449373e-30 all 2.667054e-27 likelihood   0.000543
(Tot number of parses : 21272983202438840 )

Total Rules Applied 40   Total Explicit Rules Applied 38 ( 95.00 % )
Ratio of correct rules / rules applied: 16/19

Comments: 'last week' not postmodified by 'by...'; 'la Plata' split;
'Buenos Aires' split

----------------------------------------------------------------------
Observation 4

In_II announcing_VVG the_AT award_NN1 in_II New_NP1 York_NP1 the_AT
rector_NNS1 of_IO the_AT university_NNL1 Dr_NNSB1 Nicholas_NP1
Argentato_NP1 described_VVD Mr_NNSB1 Moon_NP1 as_II a_AT1 prophet_NN1
of_IO our_APP$ time_NN1

Parsed sentence:
[V2.2
 [P1.2 [P0.3 In_II P0.3]
       [V2.1 [V1.1 [V0.12 announcing_VVG V0.12]
                   [N2.2 [DT.4 the_AT DT.4]
                         [N1.4 [N0.2 award_NN1 N0.2]
                               [P1.2 [P0.7 in_II P0.7]
                                     [N2.2 New_NP1 N2.2]P1.2]N1.4]N2.2]V1.1]
             [N0.2 York_NP1 N0.2]V2.1]P1.2]
 [V2.2
   [N2.2 [DT.4 the_AT DT.4]
         [N1.4 [N0.2 rector_NNS1 N0.2]
               [P1.2 [P0.7 of_IO P0.7]
                     [N2.2 [DT.4 the_AT DT.4
```

```
                                    [N0.2
                                     [N0.2
                                      [N0.2 [N0.2 university_NNL1 N0.2]
                                            [N0.2 Dr_NNSB1 N0.2]N0.2]
                                      [N0.2 Nicholas_NP1 N0.2]N0.2]
                                      [N0.2 Argentato_NP1 N0.2]N0.2]N2.2]P1.2]N1.4]N2.2]
   [V1.2 [V0.13 described_VVD V0.13]
         [N2.2
          [N1.4
           [N0.2 [N0.2 Mr_NNSB1 N0.2][N0.2 Moon_NP1 N0.2]N0.2]
           [P1.2 [P0.7 as_II P0.7]
                 [N2.2 [DT.4 a_AT1 DT.4]
                       [N1.4 [N0.2 prophet_NN1 N0.2]
                             [P1.2 [P0.7 of_IO P0.7]
                                   [DT.1 our_APP$ DT.1]P1.2]N1.4]N2.2]
                                                             P1.2]N1.4]
          [N0.2 time_NN1 N0.2]N2.2]V1.2]V2.2]V2.2]

Sentence length: 24 words
best 3.608663e-38 all 4.044410e-35 likelihood  0.000892
(Tot number of parses : 919495556291413934080 )

Total Rules Applied 48  Total Explicit Rules Applied 46 ( 95.83 % )
Ratio of correct rules / rules applied: 18/23

Comments: 'New York' split; no parenthetical for 'Dr..'; 'as...' too low;
'our time' split; 'of our' P1 --> P0 DT

-------------------------------------------------------------------------------
Observation 7

The_AT assembly_NN1 will_VM also_RR be_VB0 discussing_VVG the_AT
UK_NP1 immigration_NN1 laws_NN2 Hong_NP1 Kong_NP1 teenagers_NN2 in_II
the_AT church_NN1 and_CC of_RR21 church_NN1 unity_NN1 schemes_NN2

Parsed sentence:
[V2.2
 [N2.2 [DT.4 The_AT DT.4][N0.2 assembly_NN1 N0.2]N2.2]
 [V1.2 [V0.9 will_VM V0.9]
       [V1.1 [A1.4 also_RR A1.4]
             [V1.1 [V0.2 be_VB0 V0.2]
                   [V1.1 [V0.12 discussing_VVG V0.12]
                         [N2.7
                          [N2.1 [DT.3 the_AT DT.3]
                                [N1.2 [N0.2 [N0.2 UK_NP1 N0.2]
                                            [N0.2 immigration_NN1 N0.2]N0.2]
                                      [N0.1 laws_NN2 N0.1]N1.2]N2.1]
                          [N2.12
                           [N2.7
                            [N0.2 [N0.2 Hong_NP1 N0.2][N0.2 Kong_NP1 N0.2]N0.2]
                            [N1.2 [N0.1 teenagers_NN2 N0.1]
                                  [P1.2 [P0.7 in_II P0.7]
                                        [N2.2
                                         [DT.4 the_AT DT.4]
                                         [N0.2 church_NN1 N0.2]N2.2]P1.2]N1.2]N2.7]
                           [N2.12
                            [CJ.1 and_CC CJ.1]
                            [N2.7 [A0.4 of_RR21 A0.4]
                                  [N1.2 [N0.2 [N0.2 church_NN1 N0.2]
                                              [N0.2 unity_NN1 N0.2]N0.2]
                                        [N0.1 schemes_NN2 N0.1]N1.2] N2.7] N2.12]N2.12]
```

Sentence length: 21 words
best 2.152796e-31 all 3.042978e-28 likelihood  0.000707
(Tot number of parses : 1032440449833788 )

Total Rules Applied 42 Tot. Explicit Rules 38 ( 90.48 %)
Ratio of correct rules / rules applied: 20/20

Comments: correct

----------------------------------------------------------------------
Observation 9

Parsed sentence:

More_RGR important_JJ however_RR is_VBZ that_CST the_AT biblical_JJ
writers_NN2 themselves_PPX2 thought_VVD that_CST events_NN2 that_CST
followed_VVD natural_JJ laws_NN2 could_VM still_RR be_VB0 regarded_VVN
as_CSA miraculous_JJ

Parsed sentence:

[P2.1
 [A2.5
  [A2.2 [A1.5 More_RGR A1.5][A1.1 important_JJ A1.1]A2.2]
  [A1.4 [A0.4 however_RR A0.4]
        [V1.2 [V0.3 is_VBZ V0.3]
              [P1.2 [P0.3 that_CST P0.3]
                    [V2.1
                      [N2.1 [DT.3 the_AT DT.3]
                            [N1.2 [A1.1 biblical_JJ A1.1]
                                  [N0.1 writers_NN2 N0.1] N1.2] N2.1]
                      [V1.1 [N2.1 themselves_PPX2 N2.1]
                            [V0.12 thought_VVD V0.12] V1.1] V2.1] P1.2] V1.2] A1.4] A2.5]
 [P1.2 [P0.3 that_CST P0.3]
       [N1.2 [N0.1 events_NN2 N0.1]
             [P1.2 [P0.3 that_CST P0.3]
                   [V2.1 [N2.7 [V0.13 followed_VVD V0.13]
                               [N1.2 [A1.1 natural_JJ A1.1]
                                     [N0.1 laws_NN2 N0.1] N1.2] N2.7]
                         [V2.5 [V0.8 could_VM V0.8]
                               [V2.1 [A0.4 still_RR A0.4]
                                     [V1.1
                                       [V1.1 [V0.2 be_VB0 V0.2]
                                             [V1.1 [V0.12 regarded_VVN V0.12]
                                                   [P0.8 as_CSA P0.8] V1.1] V1.1]
                                       [A1.1 miraculous_JJ A1.1] V1.1] V2.1] V2.5]
                         V2.1] P1.2] N1.2] P1.2] P2.1]

Sentence length: 22 words
best 2.047509e-41 all 3.249200e-38 likelihood  0.000630
(Tot number of parses : 2341100946234064 )

Total Rules Applied 44 Total Explicit 38 (86.36%)
Ratio of correct rules / rules applied: 14/21

Comments: root P2; 'themselves thought' joined; 'thought that' split;
'as miraculous' split; rel clause not P1 + subj gap; 'still be' not V2
(explicit grammar very inadequate for this e.g.)